%% file: root.tex
\documentclass[letterpaper, 10 pt, conference]{ieeeconf}  

\IEEEoverridecommandlockouts                              

\makeatletter
\let\NAT@parse\undefined
\makeatother
\usepackage[numbers]{natbib}

\usepackage{amsmath} 
\usepackage{amssymb}  
\usepackage{bm}
\usepackage{xcolor}
\usepackage{graphicx}
\usepackage{mathtools}
\usepackage{hyperref}
\usepackage{cleveref}
\usepackage{siunitx}
\usepackage{tikz}
\usetikzlibrary{shapes}
\usepackage{multirow}
\usepackage{makecell}

\usepackage{subcaption}
\usepackage{tablefootnote}
\newcommand\copyrighttext{%
  \footnotesize \textcopyright 2023 IEEE. Personal use of this material is permitted. Permission from IEEE must be obtained for all other uses, in any current or future media, including reprinting/republishing this material for advertising or promotional purposes, creating new collective works, for resale or redistribution to servers or lists, or reuse of any copyrighted component of this work in other works.}
\newcommand\copyrightnotice{%
\begin{tikzpicture}[remember picture,overlay]
\node[anchor=south,yshift=10pt] at (current page.south) {
\setlength{\fboxsep}{0pt}%
\setlength{\fboxrule}{0pt}%
\fbox{
\parbox{\dimexpr\textwidth-\fboxsep-\fboxrule\relax}{\centering\copyrighttext}}
};
\end{tikzpicture}%
}

\Crefformat{figure}{#2Fig.~#1#3}
\Crefmultiformat{figure}{Figs.~#2#1#3}{ and~#2#1#3}{, #2#1#3}{ and~#2#1#3}

\DeclareMathOperator*{\argmin}{arg\,min}

\title{\LARGE \bf
RL + Model-based Control: Using On-demand\\Optimal Control to Learn Versatile Legged Locomotion
}

\author{Dongho Kang, Jin Cheng, Miguel Zamora, Fatemeh Zargarbashi, and Stelian Coros 
\thanks{The authors are with the Computational Robotics Lab in the Institute for Intelligent Interactive Systems (IIIS), ETH Zurich, Switzerland. %
{\tt\footnotesize \{kangd, jicheng, mimora, fzargarbashi, scoros\}@ethz.ch}}%
\thanks{We thank Zijun Hui for his assistance with the robot experiments.}
\thanks{This work has received funding from the European Research Council (ERC) under the European Union’s Horizon 2020 research and innovation programme (grant agreement No. 866480).} 
}

\begin{document}

\maketitle
\copyrightnotice

\thispagestyle{empty}
\pagestyle{empty}

\input{commands}

\begin{abstract}


This paper presents a control framework that combines model-based optimal control and reinforcement learning (RL) to achieve versatile and robust legged locomotion. 
Our approach enhances the RL training process by incorporating on-demand reference motions generated through finite-horizon optimal control, covering a broad range of velocities and gaits. 
These reference motions serve as targets for the RL policy to imitate, leading to the development of robust control policies that can be learned with reliability. 
Furthermore, by utilizing realistic simulation data that captures whole-body dynamics, RL effectively overcomes the inherent limitations in reference motions imposed by modeling simplifications. 
We validate the robustness and controllability of the RL training process within our framework through a series of experiments. 
In these experiments, our method showcases its capability to generalize reference motions and effectively handle more complex locomotion tasks that may pose challenges for the simplified model, thanks to RL’s flexibility. 
Additionally, our framework effortlessly supports the training of control policies for robots with diverse dimensions, eliminating the necessity for robot-specific adjustments in the reward function and hyperparameters.

\end{abstract}

\input{sections/introduction}
\input{sections/related_work}
\input{sections/overview}
\input{sections/reference}
\input{sections/imitation}

\input{sections/results}
\input{sections/conclusion}







\bibliographystyle{IEEEtranN}
\bibliography{root.bib}

\end{document}

%% file: commands.tex

\newcommand{\DK}[1]{{\bf\textcolor{red}{DK: #1}}}
\newcommand{\JC}[1]{{\bf\textcolor{Green}{JC: #1}}}
\newcommand{\MZ}[1]{{\bf\textcolor{blue}{MZ: #1}}}
\newcommand{\FZ}[1]{{\bf\textcolor{orange}{FZ: #1}}}
\newcommand{\Diff}[1]{{\textcolor{black}{#1}}}

\newcommand{\mboc}{MBOC}

%
\newcommand{\R}{\mathbb{R}}

\renewcommand{\r}{\mathbf{r}}
\newcommand{\f}{\mathbf{f}}
\newcommand{\uu}{\mathbf{u}}
\newcommand{\cop}{\mathbf{x}_{\text{cop}}}

\renewcommand{\v}{\mathbf{v}}

\newcommand{\X}{\mathbf{X}}
\newcommand{\U}{\mathbf{U}}
\renewcommand{\s}{\mathbf{s}}

\newcommand{\norm}[1]{\left\lVert#1\right\rVert}


\newcommand{\mycircle}{$\color{green}\pmb{\bigcirc}$}
\newcommand{\mytriangle}{$\color{orange}\pmb{\triangle}$}
\newcommand{\myx}{$\color{red}\pmb{\times}$}

%% file: sections/introduction.tex
\section{Introduction}

%

\emph{Model-based optimal control} (MBOC) and \emph{reinforcement learning} (RL) are two powerful tools widely used in legged locomotion research.
%
%
%
In \mboc{}, designers make various design choices regarding motion primitives, system dynamics, and control parameters based on simplifying assumptions \cite{wensing2022optimization}. 
This approach offers the advantage of transparency and analyzability, facilitating rapid iteration to achieve desired controller behaviors. 
However, it often constrains the controller's versatility and adaptability due to excessive reliance on assumptions and overly-simplified models.

On the other hand, RL-based approaches offer an alternative by discovering optimal control policies through trial-and-error iterations, eliminating the need for manual design choices. RL is particularly beneficial in scenarios where control and planning strategies based on simplifying assumptions are impractical \cite{lee2020learning, miki2022learning, choi2023learning}. However, the standard RL setup requires hand-crafted reward functions, which impede the induction of user-desired behaviors of a policy.

\begin{figure} 
    \vspace{0.2cm}
    \centering
    \includegraphics[width=0.95\linewidth]{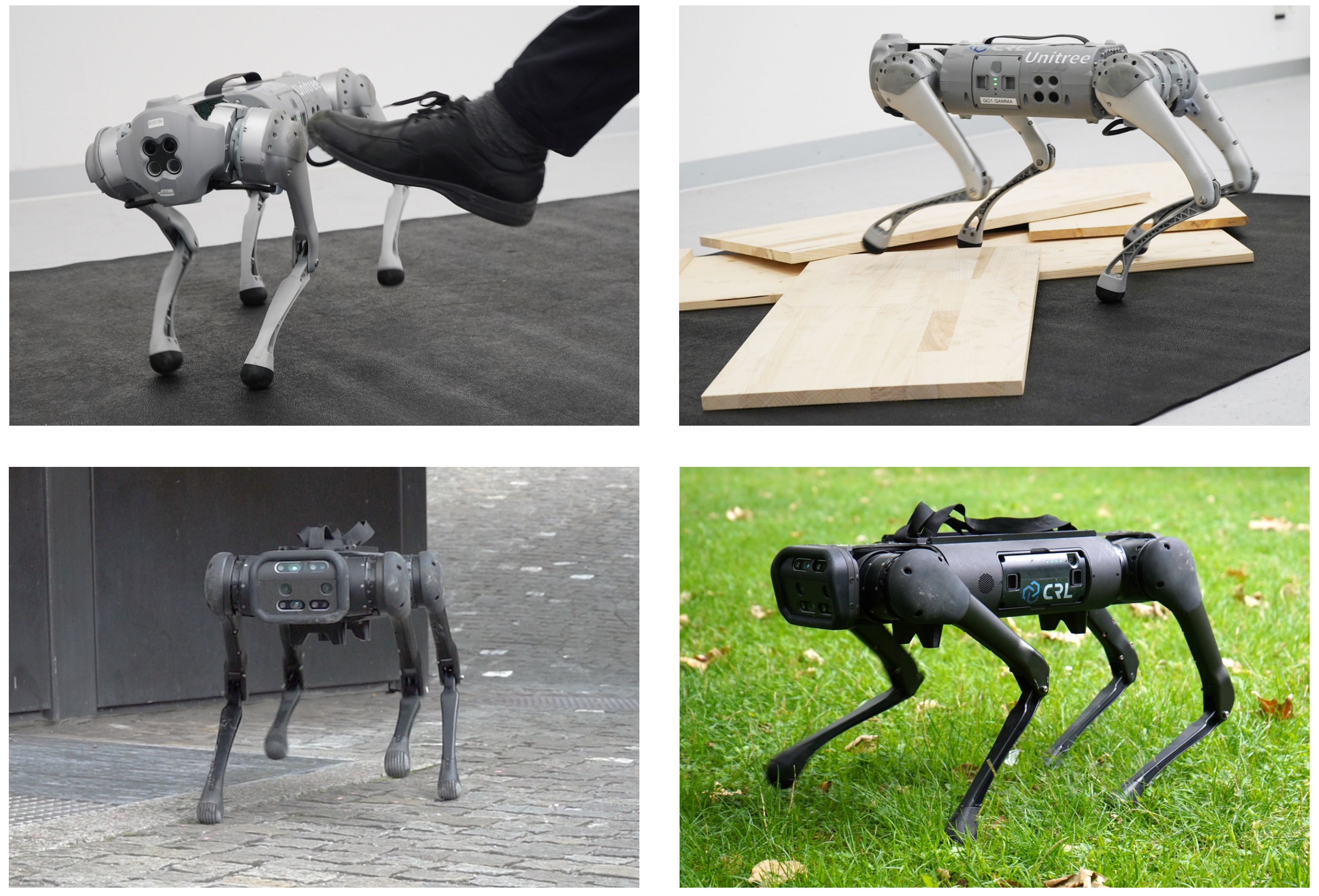}
    \caption{The snapshots of the quadruped robot \emph{Unitree Go1} (\textbf{top}) and \emph{Unitree Aliengo} (\textbf{bottom}) engaged in various locomotion tasks.} 
    \label{fig:teaser}
    \vspace{-0.6cm}
\end{figure}

In this paper, we introduce a control framework that combines \mboc{} and RL for dynamic and robust legged locomotion. 
In creating this control framework, our main goal is to determine how to complement the individual strengths of \mboc{} and RL while circumventing their intrinsic limitations. 
With this goal in mind, we structure the motion imitation problem using RL, aiming to replicate \emph{on-demand} reference motions created via an optimal control framework \cite{kang2022nonlinear, kang2022animal}.
These reference motions, covering a spectrum of gaits and target velocities, enables the training of a policy capable of generating diverse gait patterns and responding to arbitrary velocity commands. 


Our extensive evaluations highlight that our framework effectively maintains distinct gait styles, effortlessly incorporates user-defined motion parameters, and requires minimal reward shaping or additional engineering interventions. 
By tightly integrating \mboc{} into the learning process, the reliability of RL training is significantly improved, as it no longer needs to build motion skills from scratch. Furthermore, through a carefully designed imitation reward scheme, RL is able to mitigate the adverse effects of the inevitable modeling simplifications employed in our optimal control problem (OCP). Consequently, the learned policy demonstrates the ability to generalize the \mboc{} demonstrations by considering the overall impact of actions, effectively leveraging rotational dynamics, and adapting to uneven terrain. These capabilities are beyond the scope of the model employed in our OCP. \Diff{Lastly, the optimal control framework accommodates variations in the physical properties of various robots, streamlining the RL training of control policies for robots of diverse sizes without necessitating robot-specific reward or hyperparameter tuning as illustrated in \Cref{fig:teaser}.}

In summary, we introduce an RL method that incorporates the advantages of the \mboc{} approach, allowing for precise control over the behavior of a policy, while retaining the flexibility of RL to generalize to complex scenarios. Through a series of experiments involving two different quadrupedal robots, we showcase the efficacy of our approach across diverse locomotion tasks, underlining its applicability in a wide array of real-world situations.

%% file: sections/related_work.tex
\section{Related Work}



\subsection{Model-based Optimal Control for Legged Locomotion}
 
The model-based optimal control (\mboc{}) approach entails the formulation of an optimal control problem (OCP) with the goal of finding control signals that minimize an objective function, while respecting the constraints imposed by the system dynamics and physical limitations \cite{wensing2022optimization}.
In this domain, a common approach is to formulate locomotion tasks as long-horizon OCPs and utilize modern trajectory optimization (TO) techniques to reason about the results of the robot’s actions and generate complex behaviors like jumping \cite{ding2020kinodynamic, chignoli2021online}, backflipping \cite{katz_mini_2019}, and overcoming obstacles \cite{winkler2018gait, mastalli2020motion, geilinger2020computational}.

In general, the solution of a long-horizon OCP with highly nonlinear dynamics models demands substantial computational power. Model predictive control (MPC) methods, which optimize control signals in a receding horizon fashion, often utilize simplified dynamics models and relatively short time horizons to overcome this challenge. 
The inverted pendulum model \cite{xin2019online} for legged robots offers a significant reduction in the dimensionality of an OCP and facilitates the seamless integration of foothold optimization within MPC frameworks. With appropriate modifications, this model can capture dynamic motions that involve flying phases and up-and-down body motions \cite{kang2022nonlinear, kang2022animal}.
The single rigid body model (SRBM) is extensively employed in quadrupedal locomotion to effectively capture the system's rotational motions, enabling highly dynamic maneuvers involving substantial body rotation around pitch and roll axes \cite{dicarlo2018dynamic, bledt2017policy, kim2019highly, ding2019real}.

The selection of a model requires careful consideration of the trade-off between fidelity, expressiveness, and computational complexity. Achieving higher levels of realism and expressiveness comes at the cost of increased computational burdens. In our approach, which involves RL training to imitate on-demand \mboc{} demonstrations, solving OCP efficiently is crucial to reduce training time. Therefore, we adopt the low-dimensional variable-height inverted pendulum model (VHIPM) and the associated OCP formulation proposed by \citet{kang2022nonlinear, kang2022animal}. While the VHIPM may not accurately represent pitch and roll body motions or locomotion on uneven terrains, we illustrate in the subsequent sections how our motion imitation framework empowers the creation of dynamic motions encompassing body rotation and locomotion on challenging terrains.



\subsection{RL-based Legged Locomotion}


In recent years, there has been extensive research on utilizing RL for legged locomotion. RL enables the discovery of control policies by maximizing the cumulative discounted reward through trial-and-error, automating a significant portion of the control or planning strategy design process. Notably, RL can learn policies that handle modeling and environmental uncertainties by incorporating randomized training scenarios \cite{hwangbo2019learning, kumar2021rma, rudin2022learning, margolis2022rapid}. However, it should be noted that this approach often requires careful reward shaping to achieve the desired behavior, necessitating the design and balancing of auxiliary rewards to promote smooth actions, energy efficiency, and stable gait patterns.

An alternative approach is to adopt a hierarchical control structure that incorporates prior knowledge to address these challenges. 
\citet{iscen2018policies} proposed an architecture to learn a high-level RL policy that outputs parameters for the foot trajectory generator and joint signal correction. Following this paradigm, \citet{lee2020learning} and \citet{miki2022learning} utilized the RL policy to refine foot trajectories and correct footholds achieving great robustness in blind and perceptive locomotion in the wild. 


Another RL-based approach is motion imitation, where a policy is trained to replicate reference motions obtained from real-world demonstrations or other controllers. \citet{peng2018deepmimic} proposed an RL method to train control policies for simulated characters to imitate prerecorded motions from animals or humans. The training process utilizes rewards that incentivize the policy to match motions from a motion clip. This approach has enabled reproducing a wide range of behaviors without requiring excessive reward tuning. In their subsequent work, \citet{peng2020learning} successfully applied this approach to a quadruped robot using a dataset collected from dogs. 

Our approach employs the motion imitation technique introduced by \citet{peng2018deepmimic}. However, instead of real-world data, we use a reference motion generator based on \mboc{} similar to recent studies by \citet{fuchioka2022opt} and \citet{miller2023reinforcement}. This approach allows us to effectively induce desired behavior from an RL policy effectively without excessive reward shaping. Unlike these previous studies focusing on motion imitation of a single pre-generated trajectory \cite{fuchioka2022opt} or datasets of pre-generated trajectories \cite{miller2023reinforcement}, our method synthesizes reference motions on-demand during training. This strategy empowers more versatile legged locomotion control, including real-time switching of gait patterns and velocities with a single RL policy. It is worth mentioning that the previous studies by \citet{shao2022learning} and \citet{jin2022high} imitate foot trajectories generated on-demand. In contrast to their methods, we imitate \mboc{} demonstrations for both base and foot motions simultaneously. In the following chapter, we illustrate how this approach enables the development of a legged locomotion controller that is capable of producing dynamic gait patterns. \Diff{Additionally, we emphasize how leveraging \mboc{} demonstrations effortlessly simplifies the training of control policies for robots with varying physical characteristics, eliminating the need for specific adjustments tailored to each robot.}

%% file: sections/overview.tex
\begin{figure*}
    \vspace{0.2cm}
    \centering
    \includegraphics[width=0.98\linewidth]{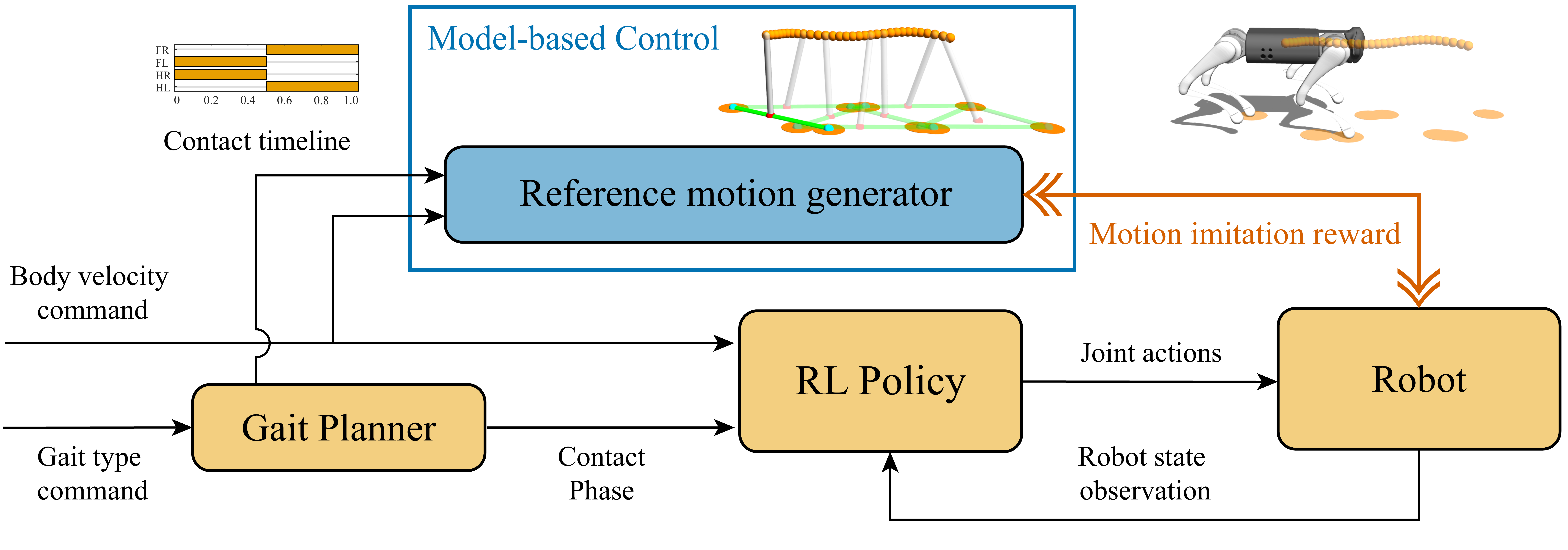}
    \caption{Overview of our framework. The objective is to train a policy that outputs joint actions to imitate the reference. The reward signal quantifies the similarity between the robot's state and reference motions generated by the MOC-based motion generator.}
    \label{fig:overview}
    \vspace{-0.5cm}
\end{figure*}

\section{Overview}

Our method combines the \mboc{} and RL approaches by training an RL policy that imitates reference robot motions generated from solving a finite-horizon OCP. 
This fusion provides a clear and intuitive means to incorporate design choices rooted in prior knowledge and desired behavior of a control policy into the process of RL training.
Furthermore, we utilize the adaptability of RL to enhance the policy's capacity for generalization across diverse and challenging scenarios, which cannot be adequately addressed solely through the simplifications inherent in the \mboc{} framework.

Our framework, depicted in \Cref{fig:overview}, develops a control policy that generates joint-level actions based on the robot's current state and high-level user commands, such as the desired body velocity and gait pattern. 
In each training episode, we randomly select a body velocity command from a range and choose a desired gait pattern from a predefined set.
The gait planner generates a contact timeline for each leg based on the desired gait pattern.
Subsequently, the reference motion generator produces a sequence of reference motions over a time horizon that follows the body velocity command and the contact timeline. These motions are created on-demand whenever new movements are required and are then stored in a queue. We employ a \mboc{} method in this stage to generate dynamically-consistent reference motions using a simplified model. Dynamically-coherent reference motions contribute to more effective RL training, particularly for motions in which dynamic effects play a crucial role. Further elaboration on this aspect is discussed in \Cref{sec:results}. 
\Diff{Additionally, it enables the training of control policies for different robots without the need for additional robot-specific reward shaping, as the variations in physical properties are already considered in the reference motions through \mboc{}.}

During the training process, an RL policy learns to map the robot's body velocity command, contact phase variables, and current state observation to joint actions by aligning the robot's state with the corresponding reference motion within a physically simulated environment. Once the policy finishes matching the entire queue of waiting reference motions, a new request for series of reference motions is sent to the motion generator. Upon completing the training process, only the gait planner and the trained RL policy are deployed to the robot hardware. 

%% file: sections/reference.tex
\section{Reference Motion Synthesis}

In the phase of reference motion generation, we solve a finite-horizon OCP based on the VHIPM \cite{kang2022nonlinear, kang2022animal} to create trajectories for the robot's base and feet. 
It is essential to maintain a low computational burden during training, as the reference motion generator is queried on-demand. Therefore, we use a low-dimensional model to minimize training time.
However, it is equally important to ensure that the model is capable of accurately capturing dynamic effects. 
The VHIPM provides a suitable compromise between model complexity and computational efficiency.

In OCP formulation, our goal is to find the system input, which includes footstep locations, a sequence of vertical accelerations, and the position of the center of pressure, that minimizes the discrepancies between the kinematic references of the base and feet motions and the predictions made by the dynamics model.
\Diff{We adopt a single shooting approach to formulate the OCP and utilize a gradient-based optimization technique to find a local minimum.}


\subsection{The Variable Height Inverted Pendulum Model}



The VHIPM represents a robot as an inverted pendulum extending from its center of pressure (CoP) to its center of mass (CoM). 
The CoP of a robot, denoted by \Diff{$\cop\in\R^3$}, is determined by a convex combination of the stance foot positions $\mathbf{s}^i\in\R^3$, which collectively define a support polygon represented by a set $\sigma$:
\vspace{0.2cm}
\begin{equation}
\label{eqn:cop}
    \cop = \sum_{\mathbf{s}^i\in\sigma} w^i \, \mathbf{s}^i,
\end{equation}
where $w^i\in\R_{\geq 0}$ is a non-negative scalar weight corresponding to $\mathbf{s}^i$ that satisfies $\sum_i w^i = 1$. 

\begin{figure} 
    \vspace{0.25cm}
    \centering
    \includegraphics[width=0.85\linewidth]{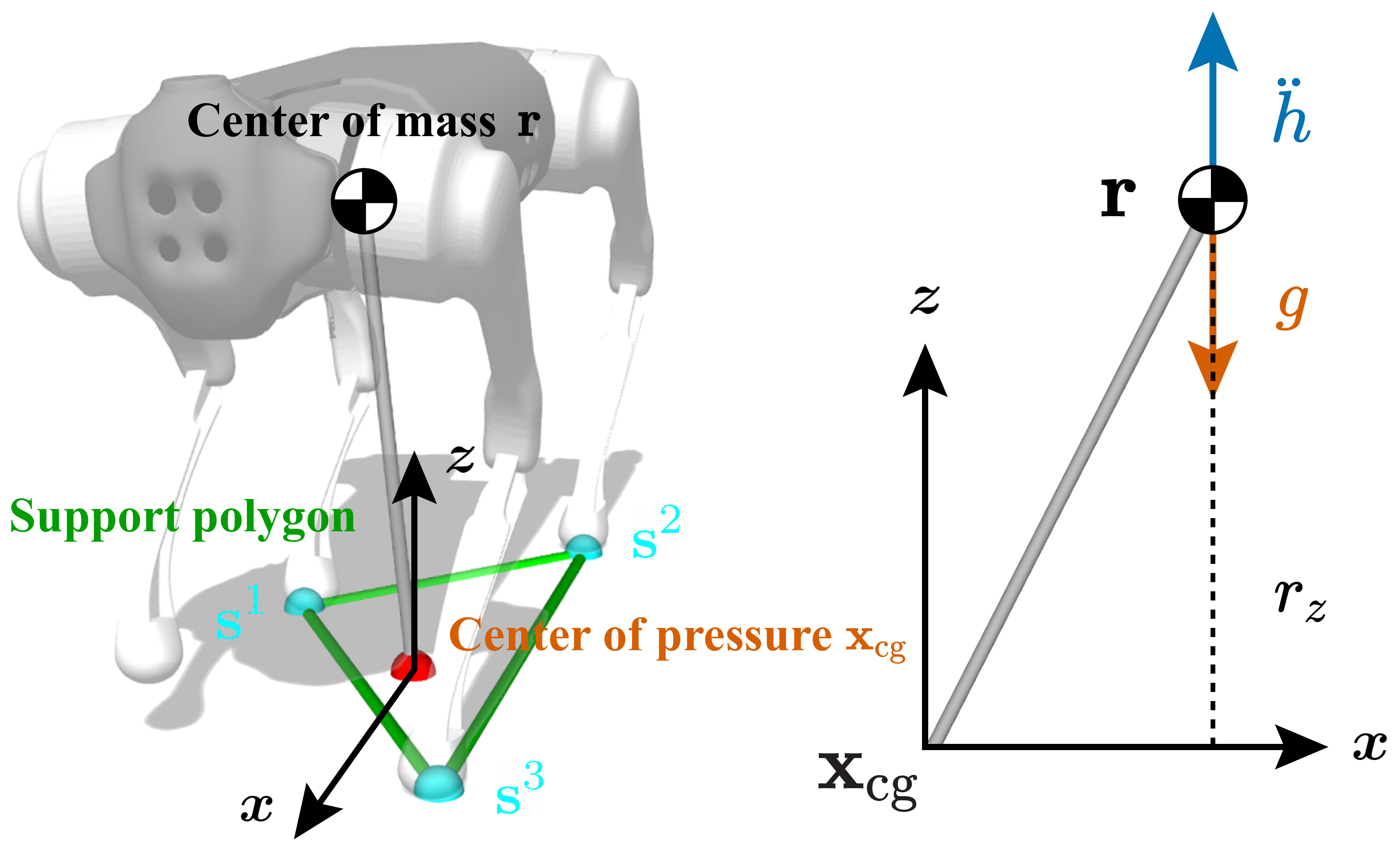}
    \caption{Quadrupedal robot \emph{Unitree Go1} represented as a variable-height inverted pendulum (\textbf{left}) and a diagram of its xz-plane projection (\textbf{right}). The VHIPM expresses a robot's CoM acceleration as a function of its CoM position, desired vertical acceleration, and CoP position.}
    \label{fig:ipm}
    \vspace{-0.4cm}
\end{figure}

Using this representation, we can express the linear acceleration of a robot's CoM denoted by $\ddot{\r}\in\R^3$ as a function of the position of the robot's CoM $\r\in\R^3$, the desired vertical acceleration $\ddot{h}\in\R$, and the CoP $\cop$ according to the following equation:
\begin{align}
    \ddot{\r} 
    &= \boldsymbol{f}_\textsc{vhipm}(\mathbf{r},\, \uu,\, \sigma) \nonumber \\
    &\coloneqq (\mathbf{r} - \cop) \frac{\ddot{h} + \|\mathbf{g}\|_2}{r_z} + \mathbf{g} \nonumber \\
    &= (\mathbf{r} - \sum_{\mathbf{s}^i\in\sigma} w^i \, \mathbf{s}^i) \frac{\ddot{h} + \|\mathbf{g}\|_2}{r_z} + \mathbf{g}. \label{eqn:ip_eom}
\end{align}
Here, we define $\uu \coloneqq \left[ \ddot{h} \; w^1 \; w^2 \; \ldots \; w^{\Vert \sigma \Vert} \right]^\top$ as the control input vector. 
For a visual representation of the VHIPM, please refer to \Cref{fig:ipm}. A detailed derivation of the equations of motion can be found in our previous work \cite{kang2022nonlinear, kang2022animal}.

It is important to highlight that the VHIPM differs from its predecessor, the linear inverted pendulum model, by considering the robot's base height as a dynamic variable rather than a constant. This allows the VHIPM to effectively capture frequent and continuous flying phases, resulting in substantial up-and-down body movements. As a result, the VHIPM plays a crucial role in generating dynamically coherent reference motions for dynamic gait patterns.

\subsection{Finite-horizon Optimal Control}

In this section, we present a concise overview of our approach to formulating and solving a finite-horizon OCP using the discretized version of the \Cref{eqn:ip_eom}.

To discretize the VHIPM, we employ a semi-implicit Euler method, resulting in the following formulation:
\begin{align}
    \r_{k+1} \hspace{-0.06cm} &\approx 2 \r_k - \r_{k-1} + \ddot{\mathbf{r}}_k \Delta t^2 \nonumber \\
    &= 2 \r_k - \r_{k-1} + \boldsymbol{f}_\textsc{vhipm}(\r_k,\,\uu_k,\,\sigma_k) \Delta t^2 \nonumber \\
    &\eqqcolon \mathbf{g}_{\textsc{vhipm},k}\left(\r_{k-1},\,\r_k,\,\uu_k,\,\sigma_k \right) \,, \label{eq:mpc-dynamics}
\end{align}
where the variables at time $t + k \, \Delta t,\, k\in\{0,\,1,\,\ldots,\,N_{\text{T}}\}$ are denoted using the subscript $k$. Here, $t$ is the current time, $\Delta t\in\R_{>0}$ is the time step duration and $N_\text{T}$ is the time horizon.

We define the stacked state $\X = \left[ \r_1 \; \r_2 \; \ldots \; \r_{N_\text{T}} \right]^\top$ and the stacked control signal $\U = \left[ \uu_0 \; \uu_2 \; \ldots \; \uu_{N_T-1} \; \s \right]^\top$, where $\s \coloneqq \left[ \mathbf{s}^1 ; \mathbf{s}^2 ; \ldots ; \mathbf{s}^{N_{\text{f}}} \right]^\top$ represents the foothold vector sorted into footfall order. Here, $N_{\text{f}}$ is the total number of footsteps over the time horizon $N_\text{T}$.

Our goal is to find an optimal system input $\U^{\star}$ that minimizes the cost function, 
\begin{equation}\label{eq:mpc-ocp}
    \U^{\star} = \argmin_{\U} \mathcal{J}(\X(\U),\,\U) \, ,
\end{equation}
where the stacked state vector $\X(\U)$ is obtained through forward integration of \Cref{eq:mpc-dynamics} given the initial conditions $\r_0$ and $\r_{-1} \coloneqq \r_0 - \v_0 \, \Delta t$. 
The cost function $\mathcal{J}(\X,\U)$, as defined in our prior work \cite{kang2022animal}, ensures that the predicted velocity and base height of the robot align with the target values. Additionally, it regularizes the footholds to avert potential kinematic singularities in the robot's legs. Moreover, it applies constraints on the input, specifically that the sum of all $w_k^i$ equals one, and each individual $w_k^i$ is greater than or equal to zero.


It should be noted that the OCP \eqref{eq:mpc-ocp} is a non-linear program due to the nonlinearity of the VHIPM model where finding a global minimum of the problem is practically infeasible. Instead, we focus on obtaining a reasonably good local minimum by employing a second-order gradient-based method \cite{kang2022nonlinear, kang2022animal, zimmermann2019optimal}. 

%% file: sections/imitation.tex
\section{Motion Imitation with Deep RL}

We train control policies using Deep RL to matches the robot's state with the reference base and foot trajectories with Proximal Policy Optimization (PPO) \cite{schulman2017proximal}. The training process involves a reward signal that measures the correspondence between the robot's state and the reference motions.


\subsection{Observation and Action Space}

The policy observes base height, gravity vector in the body frame, body linear and angular velocity, joint position and velocity, the contact phase variables in the $\sin(\cdot)$ and $\cos(\cdot)$ form, body velocity command, and previous actions. We use the contact phase parameterization proposed by \citet{shao2022learning} for each leg, which represents a swing phase as a value in $[-\pi, 0)$ and a stance phase as a value in $[0, \pi)$. This parameterization allows us to train a single policy producing diverse quadrupedal gait patterns as we demonstrate in \Cref{sec:HW}.


The action of the control policy is defined as joint target positions, which a PD controller will later take to produce target joint torques. 

\subsection{Reward Definition}

The reward function is defined as a product of the imitation reward $r^I$ and the regularizer $r^R$, namely
\begin{equation}
    r = r^I \cdot r^R.
\end{equation}
To encourage the policy to minimize the error between the current robot state and the reference motion, the imitation reward $r^I$ is defined as
\begin{equation}
    r^I = r^h \cdot r^v \cdot r^{\dot{\psi}} \cdot r^{ee},
\end{equation}
where $r^h$ minimizes the error in base height $h\in\mathbb{R}$, $r^v$ in base velocity $\mathbf{v}\in\mathbb{R}^3$, $r^{\dot{\psi}}$ in yaw rate $\dot{\psi}\in\mathbb{R}$, and $r^{ee}$ in feet positions $\mathbf{P}_{ee}\in\mathbb{R}^{4\times3}$. Notably, the feet positions $\mathbf{P}_{ee}$ are defined in the coordinate frame where the origin corresponds to the base position projected to the ground, and the orientation only retains the yaw angle of the base. This choice is made because any disturbances could potentially lead to deviation from the reference motion at a global position level.

Each reward maps the error into a scalar through the following function
\begin{equation}
    r^x = \exp\left(-\norm{\frac{\hat{x} - x}{\sigma^x}}^2\right), 
    \label{eq:exp_map}
\end{equation}
where $\hat{x}$ and $x$ are the value from the reference and the policy respectively, $\norm{\cdot}$ is the proper norm function according to the dimension of $x$, $\sigma^x$ is the sensitivity of the mapping and can either be a scalar or in the same dimension as $x$. 

The regularizer $r^R$ is defined as $r^R = r^{\Delta a} \cdot r^{\textsc{slip}} \cdot r^{\phi, \theta}$,
where $r^{\Delta a}$ regulates the action rate, $r^{\textsc{slip}}$ minimizes the contact feet velocity, and $r^{\phi, \theta}$ maintains the pitch and roll angles close to zero, thereby keeping the robot's base parallel to the ground plane. The reward function transforms the value into a scalar using a similar exponential function as \Cref{eq:exp_map}. The sensitivity values of these rewards can be found in \Cref{tab:reward-hyperparams}.

At the start of each episode, we initialize the robot's pose using a reference motion selected randomly from the reference motion queue, following the approach by \citet{peng2018deepmimic}. The episode is terminated if the robot collapses. Implementing these strategies has effectively facilitated the learning of dynamic gait patterns, particularly those involving aerial phases when the robot's feet are momentarily off the ground.

\begin{table}
    \vspace{0.25cm}
    \caption{Reward hyperparameters}
    \vspace{-0.2cm}
    \label{tab:reward-hyperparams}
    \begin{center}
    \begin{tabular}{|l|c|}
    \hline
    Reward terms $r^x$                  & Sensitivity $\sigma^x$ \\ \hline
    Base height $r^h$                   & 0.05    \\ \hline
    Base velocity$^*$ $r^v$             & [0.3, 0.1, 0.3] \\ \hline
    Base yaw rate $r^{\dot{\psi}}$      & 0.5 \\ \hline
    Feet position$^*$ $r^{ee}$          & [0.15, 0.025, 0.15] \\ \hline
    Action rate $r^{\Delta a}$          & 1.5  \\ \hline
    Feet slip $r^{\textsc{slip}}$                & 0.1  \\ \hline
    Pitch and roll $r^{\phi, \theta}$   & 0.5\\ \hline
    \end{tabular}
    \end{center}
    \caption*{$^*$ Non-scalar values are applied in forward-vertical-sideways order.}
    \vspace{-0.6cm}
\end{table}

%% file: sections/results.tex
\begin{figure*}
    \centering
    \vspace{0.05in}
    \includegraphics[width=0.98\linewidth]{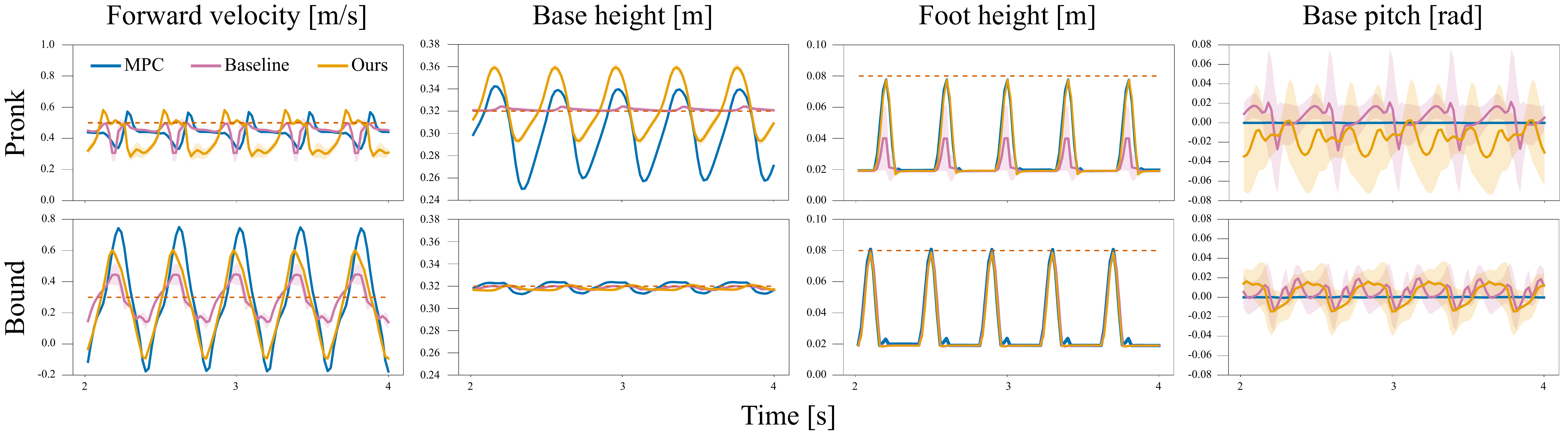}
    \caption{To observe behaviors of the MPC (in \textbf{blue}), Baseline (in \textbf{purple}) and Ours (in \textbf{yellow}) in more detail, we plot the profile of forward velocity, base height, and front-left feet trajectory for commanded velocity \SI{0.5}{\metre/\second} for pronk (\textbf{first row}) and \SI{0.3}{\metre/\second} for bound (\textbf{second row}) respectively. The colored lines are the mean of each quantity obtained from five trainings with different seeds and the shaded areas are the corresponding standard deviations. The red dotted lines stand for command velocity and motion parameters. 
    }
    \label{fig:pronk-bound}
    \vspace{-0.6cm}
\end{figure*}

\section{Results}
\label{sec:results}

We demonstrate our method through simulation and hardware experiments using the quadruped robots \emph{Unitree Go1} \cite{go1} \Diff{and \emph{Unitree Aliengo} \cite{aliengo}. The \emph{Go1} robot weighs \SI{12}{\kilogram}, while the \emph{Aliengo} robot weighs \SI{22.6}{\kilogram}.} All of the policies used in the experiments are trained on simulation data generated by an in-house tool based on the Open Dynamics Engine (ODE) \cite{smith2007open}. 
We use the PPO implementation of a RL software framework, stable-baseline3 \cite{stable-baselines3} with hyperparameters in \Cref{tab:ppo-hyperparams}.

During training, we randomly sample 3-dimensional linear and angular impulses within the ranges of $[-1.5, 1.5]\,\SI{}{\metre/\second}$ and $[-1.5, 1.5]\,\SI{}{\radian/\second}$ per dimension in the middle of each episode. 
The perturbation simulation allows us to train an RL policy capable of resisting disturbances and facilitating the transfer of the policy from simulation to the real world. 
In addition, we simulate a \SI{30}{\milli\second} actuator latency, as identified by \citet{margolis2022walktheseways}, to help bridge the sim-to-real gap. 
All RL policies used in the experiments are structured with multi-layer perceptron architecture with two fully-connected hidden layers of 256 units and the ELU activation function. The policies are queried at a rate of $\SI{50}{\hertz}$. For \emph{Go1}, the target values of base height and foot swing height are set to \SI{0.32}{\metre} and \SI{0.08}{\metre}, respectively, \Diff{while for \emph{Aliengo}, they are set to \SI{0.4}{\metre} and \SI{0.1}{\metre}.}

\subsection{Simulation Experiments}

%
%
\begin{table}
    \vspace{0.25cm}
    \caption{PPO hyperparameters}
    \vspace{-0.2cm}
    \label{tab:ppo-hyperparams}
    \begin{center}
    \begin{tabular}{|l|c||l|c|}
    \hline
    Batch size & $512$ & Discount factor & $0.95$ \\ \hline
    Number of epochs & $10$ & Learning rate & $5 \times 10^{-5}$ \\ \hline
    Value function coeff. & $0.5$ & Episode length & $128$ \\ \hline
    Entropy coeff. & $0.01$ & Initial std. deviation & $\text{exp}(-1)$ \\ 
    \hline
    \end{tabular}
    \end{center}
    \caption*{We use five random seeds of 0, 1, 10, 42, 1234 for our experiments.}
    \vspace{-0.25cm}
\end{table}

\begin{table}
\caption{The gait parameters and experiment results of control policies. A visual depiction of the gait patterns are provided in \Cref{fig:gaits}.}
\vspace{-0.2cm}
\label{tab:gaits}
\begin{center}
\begin{tabular}{|l||c|c|c|c|c|}
\hline
                            & Trot              & Pace              &   Pronk             & Bound             & Gallop                \\ \hline
Duration                    & \SI{0.5}{\second} & \SI{0.5}{\second} & \SI{0.4}{\second} & \SI{0.4}{\second} & \SI{0.5}{\second}     \\ \hline           
Duty cycle                  & 0.5               & 0.6               & 0.6               & 0.6               & 0.45                  \\ \hline
\multirowcell{3}[0pt][l]{Phase\\Offsets}    
                            & 0.5               & 0.5                   & 0             & 0                 & 0.75                  \\ 
                            & 0.5               & 0                 & 0                 & 0.5               & 0.5                   \\ 
                            & 0                 & 0.5               & 0                 & 0.5               & 0.25                  \\ \hline
                            \hline
MPC                         & \mycircle         & \mycircle         & \mycircle         & \mytriangle       & \mytriangle           \\ \hline
Baseline (Ours)             & \mycircle         & \mycircle         & \myx              & \mycircle         & \mycircle             \\ \hline
Ours                        & \mycircle         & \mycircle         & \mycircle         & \mycircle         & \mycircle             \\ \hline
\end{tabular}
\end{center}
\caption*{ \mycircle\, indicates a success over the entire range of commanded velocity, \mytriangle\, indicates instability or bad command tracking performance in some range of velocity command, and \myx\, indicates a failure to produce the target gait motion.
The duty cycle of a gait pattern is a proportion of the duration of the contact phase to a gait duration, and phase offsets are relative offsets from the phase of the front-left leg of the front-right, hind-left, and hind-right legs respectively.}
\vspace{-0.8cm}
\end{table}


In the subsequent simulation experiments, we evaluated the performance of the policy trained with our method in generating desired gait patterns, resilience against external disturbances, and traversal of challenging terrains.

\begin{figure} 
    \centering
    \includegraphics[width=\linewidth]{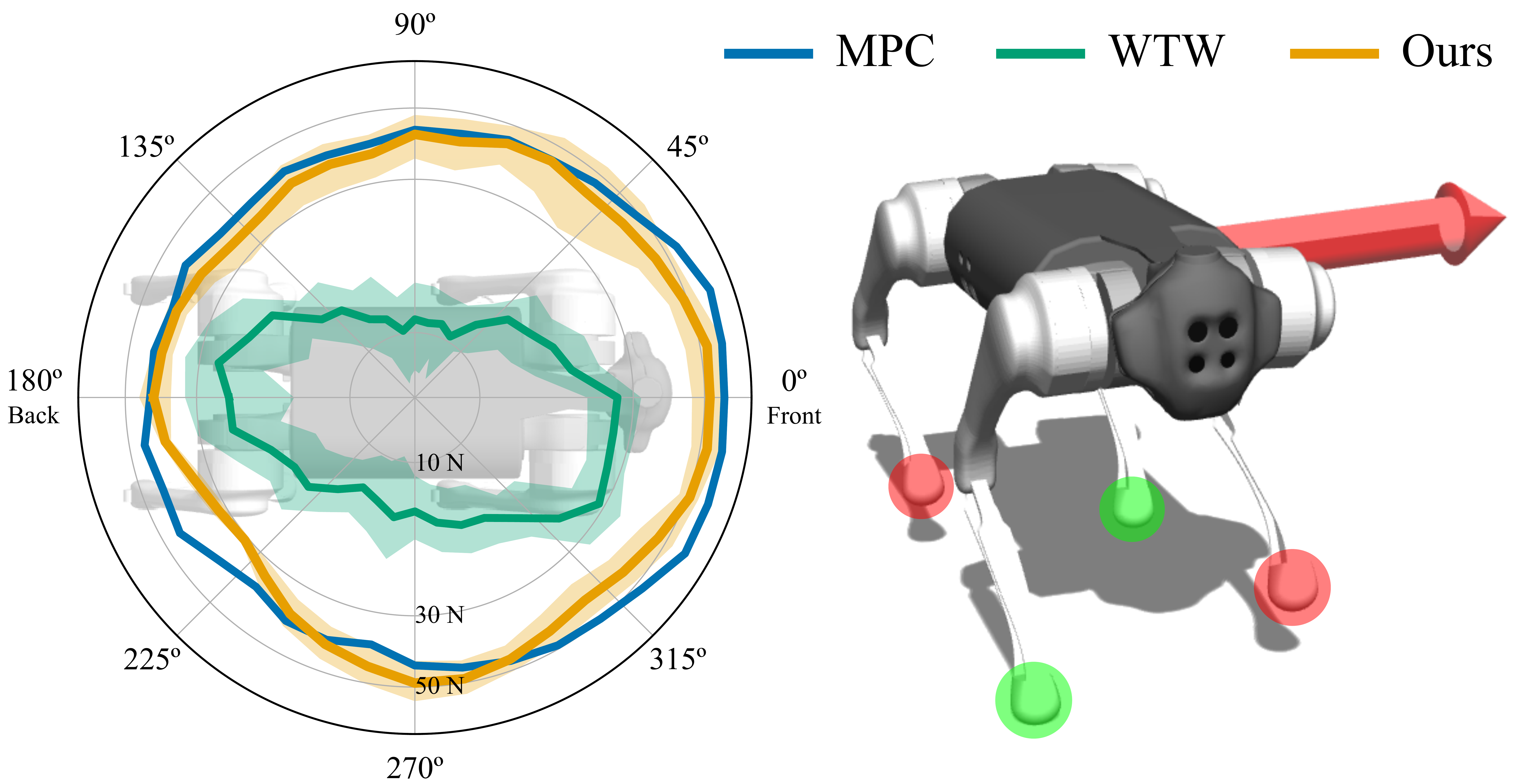}
    \caption{Snapshot of perturbation test (\textbf{right}) and the maximum pushing force along different directions that each policy withstands (\textbf{left}). The radial axis of the plot is in log scale.} 
    \label{fig:perturbation-test}
    \vspace{-0.7cm}
\end{figure}


In the first experiment, we assessed the capability of our policy to produce quadrupedal gaits with parameters in \Cref{tab:gaits}. 
We compared our approach to an MPC method proposed by \citet{kang2022nonlinear, kang2022animal} which finds a finite-horizon optimal control signal with the same simplified model and OCP formulation we use in our reference motion generator.
To convert the MPC trajectory into joint-level control signals, we combine the MPC with a whole-body control method as proposed in the previous literature \cite{kang2022animal}.

Additionally, we included a baseline policy trained using our motion imitation method with kinematic reference motions generated by numerical integration and a simple foothold planning rule [\citenum{kang2022animal}, eq. (2) and (3)], instead of the OCP solution. We introduce this policy to emphasize the importance of accounting for dynamic effects in reference motion generation when learning dynamic maneuvers.

We evaluated each policy with forward velocity commands ranging from $\SI{-0.5}{\metre/\second}$ to $\SI{1.0}{\metre/\second}$ and observe their overall behavior. 
As described in \Cref{tab:gaits}, the MPC failed to produce \emph{bound} and \emph{gallop} motions for high-velocity commands, resulting in falls. 
MPC's underlying dynamics model cannot capture the rotational dynamics resulting in zero base pitch angle all the time as illustrated in \Cref{fig:pronk-bound}.  
Rotational dynamics result in a certain level of inevitable fluctuation around the pitch axis, which is crucial for \emph{bound} and \emph{gallop} motions. 
\Diff{Notably}, our policy is trained to imitate the reference motion generated by the same dynamics model, but the RL training process can generalize the reference motions over the entire body dynamics of the robot. As a result, it learns to produce small rotational movements rather than strictly adhering to the reference motions.

Our baseline policy cannot produce \emph{pronk} motions because \emph{pronk} motion involves a momentary flying phase where the base free-falls. As the baseline policy imitates the reference motion without reflecting such dynamic effect, the policy fails to learn jumping and converges to the behavior of skating on the ground, as shown in the foot height plot. This highlights the importance of addressing the dynamic effect in generating reference motions for our motion imitation to generalize the policy over more dynamic and agile motions. 



In the second experiment, we assess the robustness of our policies in rejecting external perturbation by applying a pushing force to the robot's base in a specific direction for \SI{1}{\second} while the robot \emph{trots} in place.
In the assessment, we provide a comparison between our method with the MPC and Walk These Ways (WTW), an RL control method introduced by \citet{margolis2022walktheseways} in the assessment\footnote{To ensure a fair comparison, it is essential to match the foot contact timings for policies. We observed that the WTW policy does not always align with the commanded contact timeline. Hence, we trained the WTW policy with \emph{trot} parameters as described in \Cref{tab:gaits}, identified the resulting duty factor of 0.3, then used the same duty factor for our policy and the MPC.}. 
WTW employs a reward function that combines multiple sub-reward terms designed based on heuristics. 
Similar to our method, this method allows the specification of gait parameters and can produce all the gait patterns in \Cref{tab:gaits}.  
It should be noted that WTW used a distinct definition of observation space in their paper and trained a policy to estimate the state of a robot as well. 
On the other hand, we utilize a Kalman filter-based state estimator \cite{katz_mini_2019} for reliable estimation of base height and velocity in our hardware test. 
Hence, in the following benchmark results, we assumed that both methods have access to these quantities and used an identical definition of the observation space. 

\begin{figure} 
    \vspace{0.2cm}
    \centering
    \includegraphics[width=\linewidth]{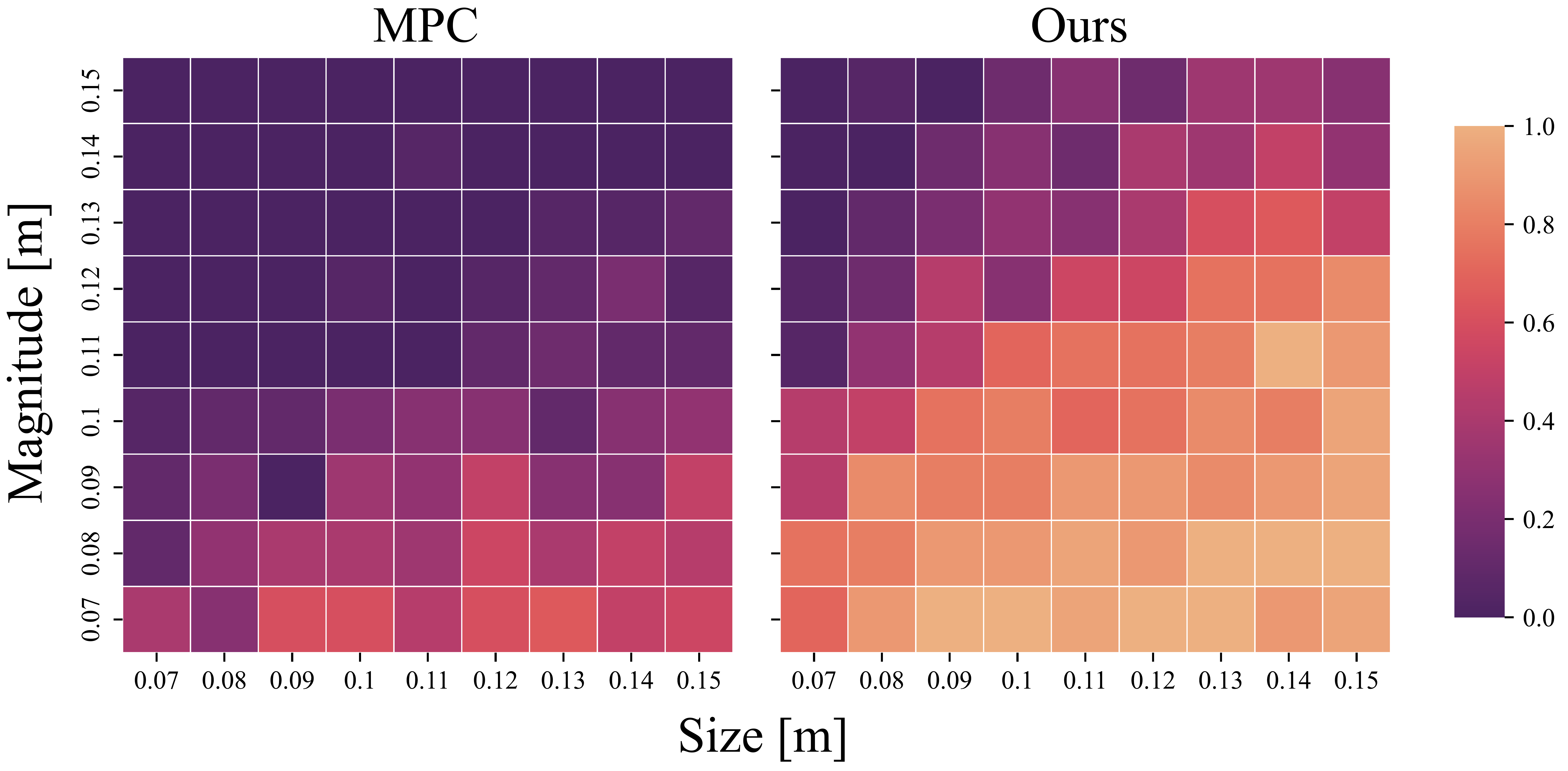}
    \caption{The success rate of the step terrain locomotion test of the MPC (\textbf{left}) and ours (\textbf{right}) with respect to different step terrain parameters. A brighter color indicates a higher success rate.} 
    \label{fig:perceptive_heatmap}
    \vspace{-0.7cm}
\end{figure}

\begin{figure*}
    \centering
    \vspace{0.05in}
    \includegraphics[width=0.98\linewidth]{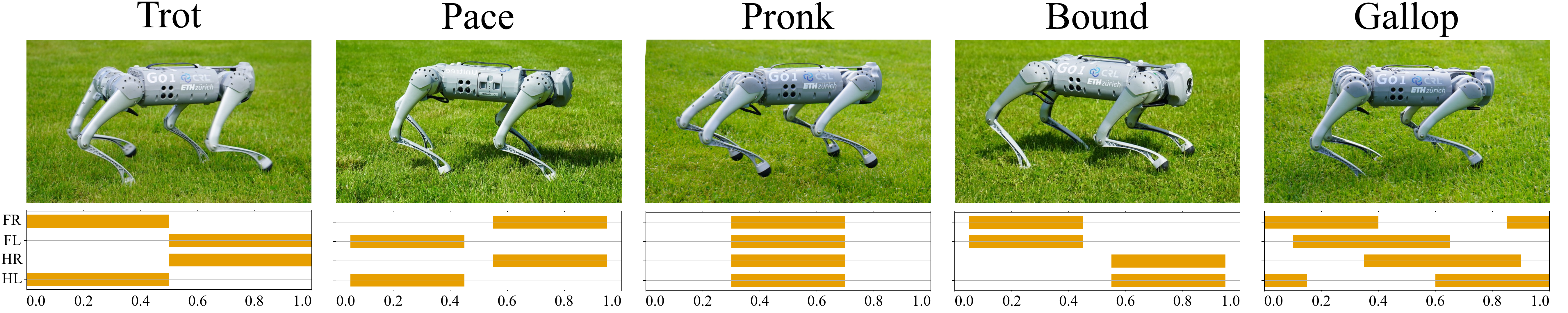}
    \caption{The snapshots of the \emph{Go1} robot (\textbf{top}) performing five different gait patterns that are graphically described in Hildebrand-style gait diagrams (\textbf{bottom}).}
    \label{fig:gaits}
    \vspace{-0.5cm}
\end{figure*}

\Cref{fig:perturbation-test} shows the maximum force magnitude along different directions that our policy, WTW policy, and the MPC can withstand without collapsing. 
In this experiment, our policy demonstrates superior performance to the WTW policy and achieves similar results as the MPC in resisting pushing forces while maintaining the \emph{trot} motion style. 
This outcome can be attributed to the robustness of our training process, which effectively preserves the walking motion styles even in the presence of random impulse perturbations. 
\Diff{We note that we could not apply the same range of impulse perturbation to the WTW policy during the training as it cannot preserve the style of motions and converges to behaviors of crouching or dragging the robot's legs consistently.}

In the final simulation experiment, we demonstrate the generalization capability of our method in handling locomotion on rough terrains, overcoming the limitations of the VHIPM in dealing with non-flat surfaces. We introduce simple modifications by using terrain measurements to adjust the reference trajectories for the base and feet, aligning them with the vertical variations of the terrain. Additionally, we include a height scan of $7 \times 11$ measurements around the base as an observation, similar to \cite{rudin2022learning}. This observation allows the robot to anticipate obstacles and react proactively, preventing conservative behaviors such as stepping in place blindly.

For the experiment, an RL policy is trained on terrains consisting of square steps of random height sampled from $[0, 0.15] \SI{}{\metre}$. The size of the square steps within an episode is uniformly sampled from $[0.07, 0.15] \SI{}{\metre}$. To evaluate the performance of the RL policy, we compare it with the modified MPC that considers the height offset of the terrain. 

The heatmaps presented in \Cref{fig:perceptive_heatmap} illustrate the success rate of 20 trials conducted on randomly generated terrains, comparing the performance of the RL policy and the MPC. The failure cases encompass collisions between the base and the ground, as well as failure to reach a boundary of $\SI{5}{\metre}$ within $\SI{10}{\second}$. Notably, our controller significantly outperforms the MPC. While the MPC struggles to navigate the terrains due to limitations in its model, our policy successfully generalizes the reference motions generated using the same model and exhibits enhanced adaptability to rough terrain locomotion. The supplementary video contains the footage of the experiment \footnote{The video is available in \url{https://youtu.be/HXwLXdOf79c}}. 

\subsection{Hardware Experiments}
\label{sec:HW}


We validated our approach by training RL policies for \emph{Go1} and \emph{Aliengo} hardware platforms to generate all the gait patterns described in \Cref{tab:gaits}. Each policy was trained using a total of \num[group-separator={,}]{16384000} data samples. \Diff{The RL training took 46 hours on the \emph{AMD Ryzen 5995WX} 64-cores CPU.}

To ensure a reliable sim-to-real transfer, we incorporated additional measures during the training phase.
Firstly, we introduced randomization of the friction coefficient of the terrain, ranging from $0.5$ to $1.25$. This variation in friction helps the policy adapt to different surface conditions and enhances its robustness.
Secondly, for each episode, we generate a random uneven terrain using Perlin noise \cite{lee2020learning} with a frequency range of $[0, 0.9]$ and a magnitude range of $[0, 0.1]\,\SI{}{\metre}$. This terrain variability allowed the policy to learn to navigate through rough surfaces. 
Thirdly, we simulated terrain inclination by randomly sampling the gravity vector within a cone of a half-angle of $\SI{10}{\deg}$ in the physics simulation. This enabled the policy to handle uneven terrains with varying slopes and further expanded its capabilities.
Finally, during training, we applied the same random impulses used for the perturbation test. This measure is crucial not only for enhancing the policy's resilience against external pushes but for achieving an effective sim-to-real transfer. The rationale lies in the following observations: as the policy converges, the robot's motions align closely with the reference motions. 
Consequently, the policy becomes more susceptible to perturbations, or potential deviations from other factors contributing to the sim-to-real gap. By applying impulses, we can generate a more diverse set of samples that assist the policy in effectively addressing the sim-to-real gap.

Our policies successfully produce diverse gait patterns while effectively withstanding external pushes and navigating rough and inclined terrains. We highlight that this robust and versatile behavior was achieved without excessive reward shaping or additional engineering. \Diff{Furthermore, we do not need any robot-specific measures as the \mboc{} demonstrations already address variance of the physical properties of different robots.} 
We provide snapshots of the hardware experiments in \Cref{fig:teaser} and \Cref{fig:gaits}. 
For a more extensive perspective, we encourage readers to view the experiment footage in our supplementary video.

%% file: sections/conclusion.tex
\section{Conclusion and Future Work}

We introduce a legged locomotion control method capable of generating diverse gait patterns while maintaining robustness against perturbations and uneven terrains. 
Our approach involves imitating dynamically coherent reference motion generated using \mboc{} with a simplified dynamics model. We further generalize these reference motions to handle more complex locomotion tasks that may challenge the simplified model, leveraging the flexibility of RL.
Our method offers an intuitive way to incorporate prior knowledge, and desired motion parameters into the RL framework, enabling the learning of versatile policies that exhibit desired behaviors without excessive reward shaping or additional engineering. 
\Diff{Moreover, our method can be readily applied to various quadruped robots without the need for robot-specific adjustments.}

To reduce the computational burden, we employ a simple VHIPM for reference motion generation. 
Notably, we have successfully extended the application of the VHIPM to handle more complex locomotion scenarios within our framework. 
Yet, our approach remains open to the integration of more advanced dynamics models. 
Importantly, we emphasize the significance of carefully considering the trade-off between strictly imitating behaviors derived from \mboc{} and allowing flexibility within RL to generalize motions. 
This trade-off will be a key aspect of our future investigations.

We are interested in extending our approach to generate a wider range of legged robot behaviors, unrestricted by the constraints of periodic or symmetric gait patterns \cite{kang2022animal, kang2021animal}. The contact phase parameterization \cite{shao2022learning} we have adopted is not limited to representing periodic or symmetric gaits. Therefore, we aim to investigate the applicability of our control method in generating more animal-like behaviors on legged robots, building upon our prior work on animal-inspired locomotion \cite{kang2022animal, kang2021animal}.
